\documentclass[]{interact}

\usepackage{epstopdf}
\usepackage[caption=false]{subfig}

\usepackage[numbers,sort&compress]{natbib}
\bibpunct[, ]{[}{]}{,}{n}{,}{,}
\renewcommand\bibfont{\fontsize{10}{12}\selectfont}
\usepackage[pagebackref=false,colorlinks=true,citecolor=blue]{hyperref}

\theoremstyle{plain}

\theoremstyle{definition}

\theoremstyle{remark}

\usepackage{graphicx}
\usepackage{url}
\usepackage{placeins}
\providecommand{\altText}[1]{}

\begin{document}

\articletype{ARTICLE}

\title{Safety-oriented sidewalk and road segmentation for smartphone-based assistive navigation}

\author{
\name{Hakan Calim\textsuperscript{1}, Anamaria Dumitrescu\textsuperscript{1,2}, Adarsh Bhandary Panambur\textsuperscript{1,3}, Huzaifa Asif\textsuperscript{1}, and Andreas Maier\textsuperscript{1}}
\affil{\textsuperscript{1}Pattern Recognition Lab, Friedrich-Alexander-University Erlangen-Nuremberg, Erlangen, Germany}
\affil{\textsuperscript{2}Department of Telecommunications, National University of Science and Technology ``Politehnica'' Bucharest, Romania}
\affil{\textsuperscript{3}Erlangen National High Performance Computing Center (NHR@FAU), Friedrich-Alexander-University Erlangen-Nuremberg, Erlangen, Germany}
}

\maketitle

\noindent\textbf{Corresponding author:} Hakan Calim, Pattern Recognition Lab, Friedrich-Alexander-University Erlangen-Nuremberg, Martensstr. 3, 91058 Erlangen, Germany. Email: \href{mailto:hakan.calim@fau.de}{hakan.calim@fau.de}
\par\medskip

\begin{abstract}
Independent sidewalk mobility is essential for blind and visually impaired pedestrians (BVIPs), yet smartphone-based assistive navigation requires perception models that distinguish walkable sidewalks from adjacent unsafe regions. This study presents a safety-oriented semantic segmentation framework for future mobile guidance. We introduce SENSATION-DS, a chest-height pedestrian-view dataset with 2,752 image-mask pairs and nine-class navigation-relevant taxonomy. External urban and sidewalk datasets were harmonized to this label space, and five segmentation architectures were evaluated using staged target-domain adaptation with mask-conditioned synthetic images and Segment Anything Model 2 (SAM2) pseudo-labels. Models were assessed using mean Intersection over Union (mIoU), road- and sidewalk-specific metrics, Road-as-Sidewalk Error Rate as a proxy false-safe measure, and Android Open Neural Network Exchange benchmarking. Synthetic augmentation generally improved segmentation accuracy, whereas SAM2 pseudo-labels more consistently reduced Road-as-Sidewalk errors. UPerNet-MobileNetV3 achieved the highest offline mIoU (0.715 $\pm$ 0.006), while DeepLabV3Plus-MobileNetV3 achieved the lowest Road-as-Sidewalk Error Rate (0.079) and highest Android runtime at 512$\times$384 (7.383 FPS). These results show that assistive sidewalk perception should be evaluated jointly by segmentation accuracy, proxy false-safe behavior, and smartphone deployment feasibility, while real-world benefit requires validation with BVIP users. This evaluation supports selecting models that balance accurate perception, conservative error behavior, and practical runtime.
\end{abstract}


\section{Introduction}

Independent mobility is fundamental for blind and visually impaired pedestrians (BVIPs), enabling access to education, employment, healthcare, and social participation \citep{ref1,ref3}. Sidewalks form the primary pedestrian corridor in urban environments but are closely connected to roads, bike lanes, crossings, curb transitions, vehicles, pedestrians, and other safety-critical scene elements. Smartphone-based assistive navigation systems must therefore identify walkable sidewalk regions while interpreting surrounding context that may indicate risk \citep{ref4,ref24}. Recent advances in deep learning-based computer vision have enabled camera-based assistive perception using smartphone sensors. Semantic segmentation is particularly suitable because it provides pixel-level understanding of sidewalks, roads, bike lanes, and surrounding objects \citep{ref6,ref10,ref11}. In assistive navigation, these masks could support estimation of the walkable corridor, detection of lateral drift, warnings for road-boundary proximity, and feedback on obstacles or uncertain road-sidewalk boundaries through audio, vibration, or wearable haptic devices. However, average segmentation accuracy alone is insufficient, as misclassifying road pixels as sidewalk may create a false-safe representation of the environment \citep{ref7,ref8}. Developing such a perception module requires addressing four key challenges: defining a segmentation task that reflects navigation risk, selecting pedestrian-view datasets with appropriate label taxonomies, expanding supervision using synthetic images or pseudo-labels without propagating boundary errors, and ensuring models remain suitable for smartphone deployment. These challenges motivate the following discussion of safety-critical segmentation errors, dataset selection, data expansion, and mobile deployment.

\subsection{Sidewalk-scene segmentation and safety-critical errors}

For assistive sidewalk navigation, semantic segmentation is valuable only when its errors are considered in relation to mobility risk. Segmentation supports tasks such as sidewalk following, obstacle awareness, and hazard warning, and has been applied to terrain awareness, sidewalk extraction, wearable and mobile perception, and assistive robotics \citep{ref6,ref7,ref8,ref9}. SideGuide further emphasizes the importance of pedestrian-view datasets for sidewalk-scene understanding \citep{ref10}. However, conventional segmentation metrics summarize overall performance and may overlook failures that are critical for BVIP navigation. For example, misclassifying a background object primarily reduces scene understanding, whereas predicting road pixels as sidewalk creates a false-safe representation of an unsafe region. This road-sidewalk confusion is therefore a safety-critical failure mode that motivates evaluation beyond overall segmentation accuracy.

\subsection{Dataset and taxonomy limitations in sidewalk navigation}

Most urban-scene segmentation benchmarks, including Cityscapes \citep{ref11}, ApolloScape \citep{ref12}, Mapillary Vistas \citep{ref13}, CamVid \citep{ref14}, and BDD100K \citep{ref15}, were developed for vehicle-centered perception. While valuable for general road-scene understanding, their viewpoint, scale, and annotation priorities do not fully reflect smartphone-based pedestrian navigation. Sidewalks often occupy small image regions and may lack accessibility-relevant annotations such as curb transitions, narrow passages, surface defects, and small obstacles. Consequently, models trained on these datasets may fail to distinguish walkable sidewalks from adjacent unsafe areas. 

Sidewalk-focused datasets reduce this viewpoint gap but remain limited. SideGuide \citep{ref10} and SANPO \citep{ref16} emphasize egocentric outdoor navigation, additional sidewalk annotations are available through Hugging Face \citep{ref17}, and the Bangladesh Footpath Dataset targets crowded pedestrian environments \citep{ref8}. However, these datasets differ in coverage, class definitions, and annotation granularity. Safety-relevant classes and scene elements such as traffic light, traffic sign, bicycle, vehicle/car, and person may be omitted or merged into broader categories, limiting transferability and hazard-specific reasoning. A safety-oriented sidewalk segmentation framework therefore requires both pedestrian-centric imagery and a label taxonomy that distinguishes walkable regions, unsafe surfaces, and surrounding hazards.

\subsection{Data expansion through synthetic images and pseudo-labels}

Even with pedestrian-relevant imagery and a navigation-oriented label space, available training data may not capture the full variability of sidewalk environments, particularly around road-sidewalk boundaries, curb regions, occlusions, and diverse urban scenes. Diffusion-based image synthesis can increase visual diversity by generating images conditioned on semantic layouts \citep{ref18}, while ControlNet improves structural consistency during generation \citep{ref19}. This allows road-sidewalk geometry to be preserved while varying image appearance. However, synthetic data may introduce unrealistic textures, inaccurate boundaries, or inconsistent image--label relationships, requiring empirical evaluation of its usefulness.

Pixel-accurate annotation is also expensive at pedestrian scale. Foundation segmentation models can alleviate this burden by generating pseudo-labels for additional real-world images. Segment Anything Model 2 (SAM2) provides promptable image and video segmentation with improved temporal mask propagation and has demonstrated competitive performance in road-obstacle and related tasks \citep{ref20,ref21}. Nevertheless, studies on long-tailed work-zone scenes highlight the need for targeted data and target-domain adaptation \citep{ref22}. Pseudo-labels may contain boundary errors, class ambiguities, missed small objects, and domain-shift failures, particularly near sidewalk, road, and hazard boundaries that are critical for safe navigation.

\subsection{Mobile deployment and user-centered relevance}

Assistive navigation requires more than high offline segmentation accuracy. Smartphone-based systems must provide timely feedback under constraints such as limited computational resources, battery capacity, and changing outdoor conditions. Previous work has explored real-time smartphone navigation \citep{ref23,ref24}, lightweight segmentation for on-device deployment \citep{ref8}, and reports persistent reliability and usability challenges for BVIP navigation applications \citep{ref5}. Segmentation models should therefore be evaluated not only by accuracy but also by their suitability for mobile deployment.

Existing studies address individual aspects of this problem, including mobile navigation \citep{ref23,ref24}, pedestrian-view datasets with mobile-oriented baselines \citep{ref16}, and foundation models for weak supervision \citep{ref45}. However, fewer studies integrate pedestrian-view dataset alignment, safety-oriented road-sidewalk error analysis, controlled data expansion, pseudo-label supervision, and smartphone deployment within a unified framework. Such integration is essential because models for BVIP navigation must accurately distinguish walkable sidewalks from unsafe road regions while remaining suitable for real-time mobile use.

User-centered evaluation is equally important. Assistive navigation systems should reflect the strategies, preferences, and risk tolerance of BVIPs, and participatory design can improve task definition, label selection, and interpretation of segmentation errors \citep{ref46,ref47}. In sidewalk-scene segmentation, even small boundary errors may become safety-critical if they incorrectly represent road regions as walkable.

Together, these considerations motivate a safety-oriented evaluation framework that combines deep learning-based segmentation, pedestrian-relevant label alignment, controlled data expansion, road-sidewalk error analysis, and smartphone feasibility.

\section{Objectives}

This study investigates safety-oriented semantic segmentation as a perception module for smartphone-based assistive navigation for BVIPs. The primary objective is to determine how sidewalk-scene segmentation models should be trained, evaluated, and selected when their outputs may support future mobile guidance in safety-critical pedestrian environments. We define Road-as-Sidewalk prediction as a false-safe failure mode because it represents unsafe road regions as walkable. Accordingly, models are evaluated using both conventional segmentation metrics and road-sidewalk-specific safety measures. To provide a consistent target domain, we introduce SENSATION-DS, a chest-height pedestrian-view dataset with a navigation-oriented taxonomy distinguishing sidewalks, roads, bike lanes, and surrounding hazard-related classes.

The experimental objectives are to:
\begin{enumerate}
\item harmonize external urban and sidewalk-scene datasets to the SENSATION-DS taxonomy and evaluate their contribution to transfer learning;
\item assess staged target-domain adaptation across lightweight and compact semantic segmentation architectures;
\item determine whether mask-conditioned synthetic images and SAM2-generated pseudo-labels improve training; and
\item export selected models to Open Neural Network Exchange (ONNX) and benchmark them in an Android application to evaluate practical smartphone deployment.
\end{enumerate}

Dataset design, class selection, and interpretation of safety-critical errors are informed by the experience of a blind first author, recognizing that the practical impact of segmentation errors depends on mobility context. A visually small road-sidewalk boundary error, for example, may become safety-critical if it incorrectly represents a road region as walkable. The study therefore evaluates both segmentation performance and mobile deployment as a step toward future user-centered assistive navigation systems.

\section{Methods}

\begin{figure}[!t]
\centering
\includegraphics[width=0.98\textwidth]{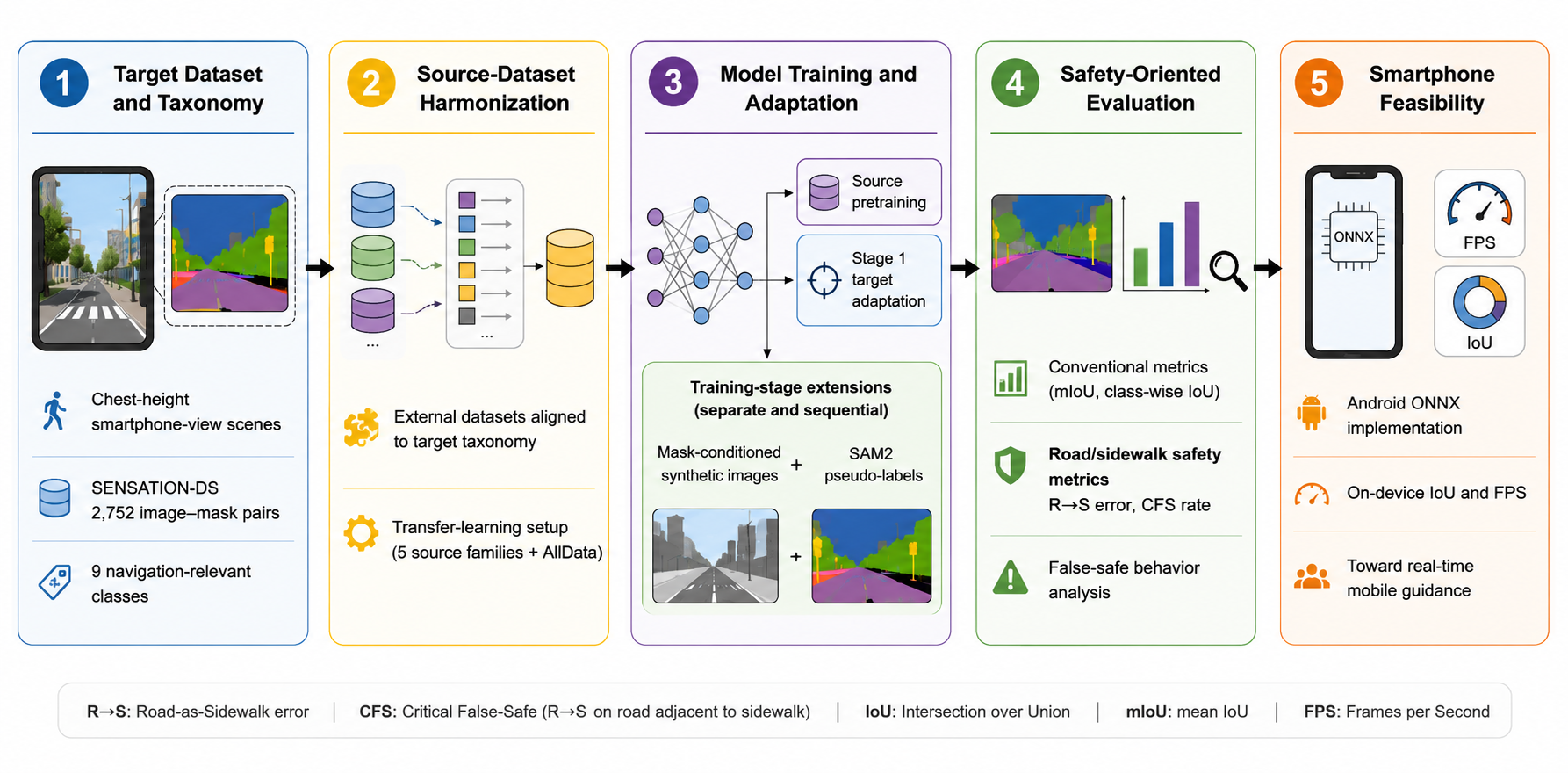}
\caption{\textbf{Overview of the proposed safety-oriented sidewalk segmentation framework for smartphone-based assistive navigation.} The workflow consists of five stages: construction of the SENSATION-DS target dataset and navigation-relevant taxonomy; harmonization of external source-data families for transfer learning; model training with staged target-domain adaptation and controlled synthetic and SAM2 pseudo-label supervision extensions \citep{ref20}; offline evaluation using conventional segmentation metrics and road-sidewalk-specific false-safe measures; and Android ONNX benchmarking for smartphone deployment feasibility.}
\altText{Five-stage horizontal workflow diagram for safety-oriented sidewalk segmentation. Stage~1, Target Dataset and Taxonomy, shows chest-height smartphone-view scenes, SENSATION-DS with 2,752 image-mask pairs, and nine navigation-relevant classes. Stage~2, Source-Data Harmonization, shows external datasets aligned to the target taxonomy and prepared for transfer learning across five source-data families and AllData. Stage~3, Model Training and Adaptation, shows source-data pretraining, Stage~1 target adaptation, and training-stage extensions using mask-conditioned synthetic images and SAM2 pseudo-labels. Stage~4, Safety-Oriented Evaluation, shows conventional segmentation metrics such as mean Intersection over Union (mIoU) and class-wise Intersection over Union (IoU), road/sidewalk safety metrics including Road-as-Sidewalk Error Rate and Critical False Safe Rate, and false-safe behavior analysis. Stage~5, Smartphone Feasibility, shows Android ONNX implementation, on-device IoU and frames per second (FPS) evaluation, and the transition toward real-time mobile guidance. Arrows connect the stages from dataset construction to deployment feasibility.}
\label{fig:framework_overview}
\end{figure}

Figure~\ref{fig:framework_overview} summarizes the workflow used to evaluate safety-oriented semantic segmentation for smartphone-based assistive navigation. The study proceeds from construction of the SENSATION-DS target dataset and harmonization of external source-data families, through staged target-domain adaptation with synthetic and SAM2 pseudo-label supervision extensions, to offline safety-oriented evaluation and Android ONNX deployment benchmarking. This structure aligns the experimental pipeline with the central objective of the study: selecting segmentation models not only by global accuracy, but also by road-sidewalk false-safe behavior and practical mobile feasibility.

\subsection{SENSATION-DS Dataset and Navigation-Relevant Taxonomy}

SENSATION-DS was defined as the target dataset for sidewalk-scene segmentation from a chest-height pedestrian viewpoint, reflecting the intended smartphone-based navigation setting. It adopts a 9-class taxonomy comprising background, road, sidewalk, bike lane, person, vehicle/car, bicycle, traffic light, and obstacle to represent both walkable space and surrounding context. Road and sidewalk are the navigation-critical classes and correspond to class~1 and class~2, respectively. Dataset design, class selection, and interpretation of safety-critical failures were informed by the experience of a blind first author.

The dataset contains 2,752 image-mask pairs assembled from three complementary sources: pedestrian-view samples from Mapillary Vistas \citep{ref13}, sidewalk image-mask pairs released through Hugging Face \citep{ref17}, and newly captured chest-height images from German urban environments annotated using the EXACT toolset \citep{ref26}. A fixed train/validation/test split of 2,190/281/281 images (approximately 80\%/10\%/10\%) was used throughout the study. The validation set supported training-stage comparison and model selection, while the held-out test set was reserved for final evaluation.

\subsection{Source-Data Harmonization and Transfer Setup}

To evaluate transfer learning under a common pedestrian-navigation label space, we used five source-data families: ApolloScape \citep{ref12}, Cityscapes \citep{ref11}, SANPO \citep{ref16}, SideGuide \citep{ref10}, and AllData, the combined multi-source setting. As these datasets differ in viewpoint, scene composition, annotation detail, and class definitions, all labels were mapped to the SENSATION-DS 9-class taxonomy using dataset-specific mapping tables before training. Classes that could not be mapped reliably were assigned to an ignore label and excluded from loss computation and metric evaluation. This harmonization enabled direct comparison of models across all source-data families under a consistent target taxonomy.

\subsection{Model Architectures, Optimization, and Staged Target-Domain Adaptation}

We evaluated five semantic segmentation architectures implemented with \texttt{segmentation\_models\_pytorch} \citep{ref29} and \texttt{timm} encoders \citep{ref30}: DeepLabV3Plus \citep{ref31}, Feature Pyramid Network (FPN) \citep{ref32}, Pyramid Attention Network (PAN) \citep{ref33}, UPerNet \citep{ref34}, and SegFormer \citep{ref35}. DeepLabV3Plus, FPN, PAN, and UPerNet used ImageNet-pretrained MobileNetV3-Large-0.75 encoders \citep{ref28,ref36}, while SegFormer used the MiT-B0 encoder \citep{ref35}. These architectures were selected to compare lightweight convolutional encoder--decoder models with a compact transformer while remaining suitable for smartphone deployment.

\begin{figure}[!t]
\centering
\includegraphics[width=0.92\textwidth]{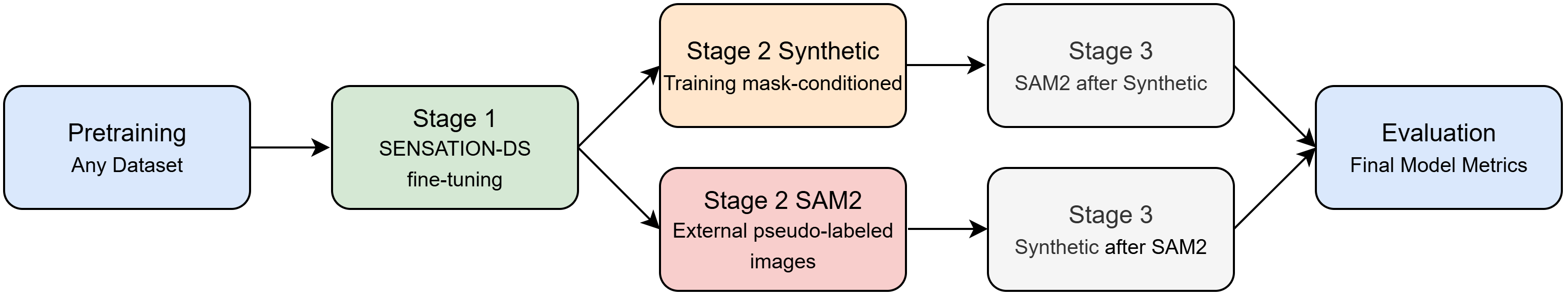}
\caption{\textbf{Training-stage pipeline for staged target-domain adaptation to SENSATION-DS.} Models are first pretrained on a source-data family and then fine-tuned on the SENSATION-DS training split. Two controlled Stage~2 extensions are evaluated separately: mask-conditioned synthetic images and SAM2 pseudo-labeled external images \citep{ref20}. In Stage~3, the two extensions are applied sequentially in both orders to compare whether Synthetic$\rightarrow$SAM2 or SAM2$\rightarrow$Synthetic refinement better supports validation performance and safety-oriented road-sidewalk behavior.}
\altText{Horizontal workflow diagram showing staged target-domain adaptation. The process begins with source-data pretraining, followed by Stage~1 fine-tuning on SENSATION-DS. The pipeline then splits into two Stage~2 branches: one using mask-conditioned synthetic images and one using SAM2 pseudo-labeled external images. Each branch is followed by a Stage~3 cross-enhancement step, applying the two extensions in the opposite order. Both branches then lead to evaluation of final model metrics.}
\label{fig:training_stage_pipeline}
\end{figure}

The staged target-domain adaptation pipeline is illustrated in Figure~\ref{fig:training_stage_pipeline}. Models were pretrained on each source-data family, fine-tuned on the SENSATION-DS training split (Stage~1), extended independently with synthetic images or SAM2 pseudo-labeled images (Stage~2), and finally trained using both extensions in the two possible sequential orders (Stage~3). This design isolates the contributions of source pretraining, target-domain adaptation, and the two supervision extensions.

Unless otherwise stated, all models shared the same optimization and augmentation protocol. Training used AdamW \citep{ref38} with a weight decay of \(1\times10^{-4}\), a OneCycle learning-rate schedule \citep{ref39}, and Albumentations-based augmentation \citep{ref27}. The supervised objective combined Dice loss and cross-entropy loss with validation-selected weights of 0.3 and 0.7, respectively. Ignore-label pixels were excluded from both loss and metric computation. Each configuration was repeated across five random seeds where available, and results are reported as mean \(\pm\) standard deviation.

\subsection{Controlled Training Extensions with Synthetic Images and SAM2 Pseudo-Labels}

Two controlled training-stage extensions were evaluated to increase supervision beyond the SENSATION-DS training split. The first used mask-conditioned synthetic images to increase appearance diversity while preserving the semantic layout, whereas the second used SAM2-generated pseudo-labels to incorporate additional real-world sidewalk scenes. Both extensions were applied only during training and excluded from validation and the held-out test set.

\begin{figure}[!t]
\centering
\includegraphics[width=0.98\textwidth]{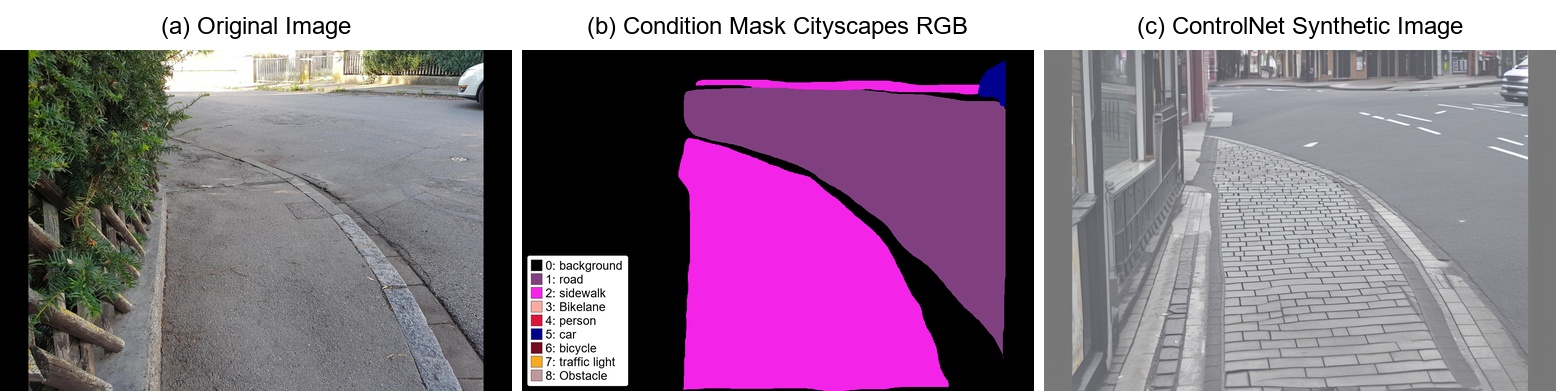}
\vspace{0.5em}

\includegraphics[width=0.98\textwidth]{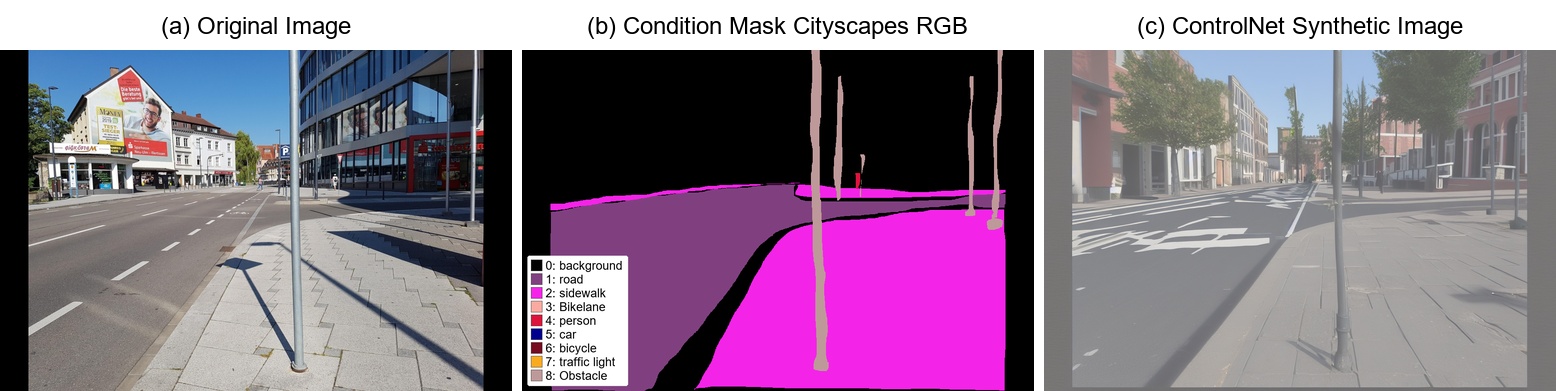}
\vspace{0.5em}

\includegraphics[width=0.98\textwidth]{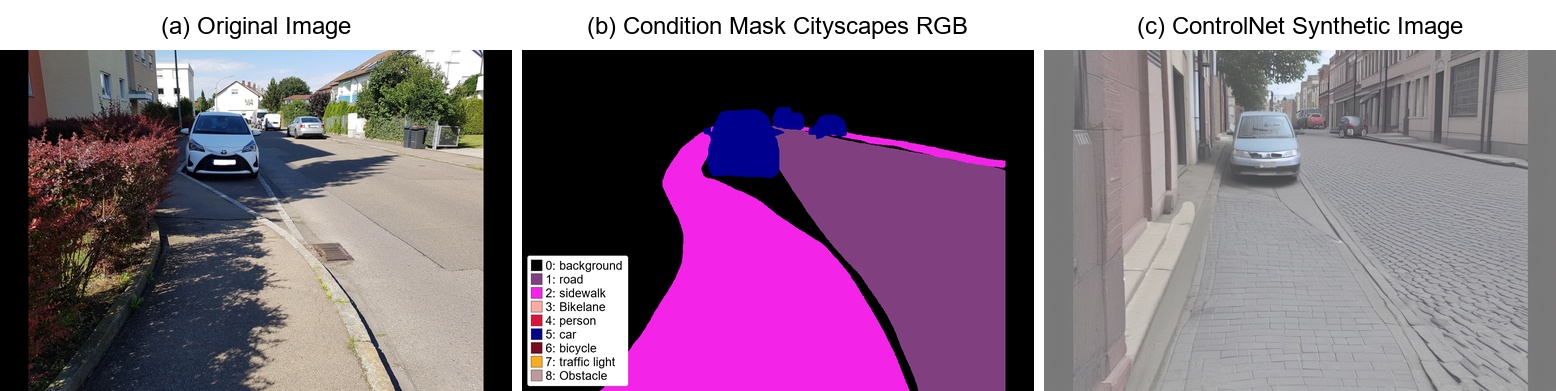}
\caption{\textbf{Accepted ControlNet-generated synthetic sidewalk-scene examples used for controlled training-stage augmentation.} Each row shows the source image, semantic condition mask, and corresponding ControlNet-generated image \citep{ref11,ref19}. The retained examples illustrate different road-sidewalk geometries and were used exclusively for training augmentation.}
\altText{Three-row qualitative figure showing accepted ControlNet synthetic training examples. In each row, the left panel is the original source image, the middle panel is the Cityscapes-style semantic condition mask, and the right panel is the ControlNet-generated synthetic image. The first row shows a curved sidewalk next to a road or paved surface; the condition mask separates sidewalk, road, background, and a small vehicle/car area; the synthetic output preserves the curved sidewalk boundary while changing the surrounding urban appearance. The second row shows a broad urban road and sidewalk with buildings, crosswalk markings, and sign or pole structures; the condition mask represents a large sidewalk region, adjacent road, and vertical pole-like objects; the synthetic output retains the street layout and generates a different city-street appearance. The third row shows a residential street with parked vehicles and a sidewalk edge; the mask marks a curving sidewalk, road, and vehicle/car regions; the synthetic image preserves the road-sidewalk geometry while changing texture, building facades, and vehicle appearance. The figure demonstrates that accepted synthetic samples maintain the semantic layout needed for training augmentation while varying visual appearance.}
\label{fig:controlnet_synthetic_examples}
\end{figure}

Mask-conditioned synthetic images were generated from SENSATION-DS training masks using diffusion-based semantic image synthesis with ControlNet-style conditioning \citep{ref18,ref19}. This approach varied scene appearance while preserving the underlying semantic structure. Synthetic samples were filtered before use, and only accepted images were included in the corresponding fine-tuning stage. Representative examples are shown in Figure~\ref{fig:controlnet_synthetic_examples}.

For pseudo-label supervision, SAM2 \citep{ref20} generated segmentation masks for additional unlabeled chest-height sidewalk images collected in German urban environments. These images were used only for training, excluded from the SENSATION-DS validation set and held-out test set, and anonymized where appropriate.

The two extensions address complementary limitations of the training data: synthetic images increase appearance diversity while preserving semantic structure, whereas SAM2 pseudo-labels introduce additional real-world scenes with automatically generated annotations. Restricting both to the training stage prevents data leakage and enables controlled evaluation of their impact on segmentation performance and safety-oriented road-sidewalk behavior.

\subsection{Safety-Oriented Evaluation and Smartphone Benchmarking}

Offline evaluation used the validation set for training-stage comparison and candidate selection, while the held-out test set was reserved for final reporting. Performance was assessed using mean Intersection over Union (mIoU) and class-wise Intersection over Union (IoU) for the road and sidewalk classes. Invalid, unmapped, and ignore-label pixels were excluded from both loss computation and metric evaluation. Safety-oriented evaluation included Road-as-Sidewalk Error Rate as the primary false-safe metric, together with Critical False Safe Rate, Sidewalk-as-Road Error Rate, Safe-Walkable Precision, Safe-Walkable Recall, Navigational Risk Error, confusion matrices, and per-image failure summaries.

Smartphone feasibility was evaluated using ONNX-exported models in the SENSATION Android application on Snapdragon-class devices \citep{ref23}. The benchmark reports on-device IoU and frames per second (FPS), enabling segmentation performance and runtime to be assessed together. Results are reported as mean \(\pm\) standard deviation across five random seeds where available.

\FloatBarrier

\section{Results}

The results are reported in three steps: family-wise validation of staged target-domain adaptation, offline comparison of selected candidates using accuracy and safety-oriented road/sidewalk metrics, and Android ONNX benchmarking for smartphone feasibility. Validation results were used for training-stage comparison and candidate selection, while the held-out test set was reserved for final reporting of selected candidates.

\subsection{Effect of Source-Data Family and Staged Target-Domain Adaptation}

Table~\ref{tab:family_stage_summary} summarizes validation performance across source-data families and training stages. The target-only SENSATION-DS baseline was stable across comparisons, with mIoU values around 0.683 and road/sidewalk IoU values around 0.738 and 0.857, respectively. This provided a consistent reference for evaluating transfer learning and staged refinement.

Direct source-to-target adaptation alone was not beneficial. The source$\rightarrow$target setting reduced mIoU for all source-data families, with the largest decreases observed for SideGuide, from 0.6829 to 0.4057, and ApolloScape, from 0.6831 to 0.5039. Although some Stage~1 settings showed lower false-safe rates, these changes occurred with substantial losses in overall, road, and sidewalk segmentation performance and were therefore not interpreted as meaningful safety improvements.

Controlled training extensions improved performance more consistently. Synthetic augmentation increased mIoU for AllData, ApolloScape, Cityscapes, and SANPO, with the largest gain for Cityscapes, from 0.6832 to 0.7038. It also improved Road IoU across all source-data families. Adding SAM2 pseudo-labels produced smaller and less consistent mIoU gains, but reduced Road-as-Sidewalk Error Rate in all Stage~2 source-data family comparisons; for example, Cityscapes decreased from 0.1022 to 0.0857 and SANPO from 0.1017 to 0.0848.

The Stage~3 ordering further separated accuracy-oriented and safety-oriented outcomes. SAM2$\rightarrow$Synthetic produced the highest mIoU for most source-data families, including Cityscapes at 0.7064 and AllData at 0.7021. In contrast, Synthetic$\rightarrow$SAM2 produced the lowest Road-as-Sidewalk Error Rate in every source-data family comparison, with values of 0.0838 for AllData, 0.0916 for ApolloScape, 0.0808 for Cityscapes, 0.0830 for SANPO, and 0.0902 for SideGuide. These findings support the central premise of the study: for assistive sidewalk navigation, staged target-domain adaptation should be interpreted using both global segmentation accuracy and road/sidewalk-specific false-safe metrics.

\begin{table*}[!t]
\centering
\caption{\textbf{Family-wise summary of staged SENSATION-DS target-domain adaptation.} Values in parentheses denote changes relative to the corresponding target-only SENSATION-DS baseline for each source-data family. R$\rightarrow$S denotes Road-as-Sidewalk Error Rate and CFS denotes Critical False Safe Rate. Higher is better for mIoU, Road IoU, and Sidewalk IoU; lower is better for R$\rightarrow$S and CFS. Bold values indicate the best result for each metric within each source-data family comparison block.}
\altText{Large validation table comparing staged SENSATION-DS target-domain adaptation across five source-data families: AllData, ApolloScape, Cityscapes, SANPO, and SideGuide. For each family, rows report target-only baseline, source-to-target only, synthetic-image extension, SAM2 pseudo-label extension, SAM2$\rightarrow$Synthetic ordering, and Synthetic$\rightarrow$SAM2 ordering. Columns report mIoU, Road IoU, Sidewalk IoU, Road-as-Sidewalk Error Rate, and Critical False Safe Rate, with parenthetical changes relative to the corresponding target-only baseline. Higher values are better for mIoU and IoU columns, while lower values are better for Road-as-Sidewalk Error Rate and Critical False Safe Rate. The table shows that the synthetic-image extension and SAM2$\rightarrow$Synthetic ordering often improve mIoU, while Synthetic$\rightarrow$SAM2 gives the lowest Road-as-Sidewalk Error Rate in each source-data family block.}
\label{tab:family_stage_summary}
\scriptsize
\setlength{\tabcolsep}{2.5pt}
\resizebox{\textwidth}{!}{%
\begin{tabular}{llccccc}
\toprule
Source-data family & Training setting & mIoU $\uparrow$ & Road IoU $\uparrow$ & Sidewalk IoU $\uparrow$ & R$\rightarrow$S $\downarrow$ & CFS $\downarrow$ \\
\midrule

AllData & Target-only baseline & 0.6832 & 0.7377 & 0.8572 & 0.1022 & 0.0616 \\
AllData & source$\rightarrow$target only & 0.5480 (-0.1352) & 0.4834 (-0.2543) & 0.5064 (-0.3508) & 0.0869 (-0.0153) & \textbf{0.0516} (-0.0100) \\
AllData & + Synthetic images & 0.6977 (+0.0145) & \textbf{0.7643} (+0.0265) & \textbf{0.8638} (+0.0066) & 0.0900 (-0.0122) & 0.0565 (-0.0052) \\
AllData & + SAM2 pseudo-labels & 0.6924 (+0.0092) & 0.7461 (+0.0083) & 0.8565 (-0.0008) & 0.0900 (-0.0122) & 0.0566 (-0.0050) \\
AllData & SAM2$\rightarrow$Synthetic & \textbf{0.7021} (+0.0189) & 0.7606 (+0.0228) & 0.8633 (+0.0061) & 0.0908 (-0.0114) & 0.0569 (-0.0047) \\
AllData & Synthetic$\rightarrow$SAM2 & 0.6985 (+0.0153) & 0.7576 (+0.0199) & 0.8602 (+0.0030) & \textbf{0.0838} (-0.0184) & 0.0540 (-0.0076) \\

\midrule
ApolloScape & Target-only baseline & 0.6831 & 0.7380 & \textbf{0.8572} & 0.1016 & 0.0613 \\
ApolloScape & source$\rightarrow$target only & 0.5039 (-0.1792) & 0.4846 (-0.2534) & 0.5343 (-0.3229) & 0.1205 (+0.0188) & 0.0721 (+0.0108) \\
ApolloScape & + Synthetic images & 0.6903 (+0.0072) & \textbf{0.7466} (+0.0087) & 0.8560 (-0.0011) & 0.0976 (-0.0040) & 0.0619 (+0.0006) \\
ApolloScape & + SAM2 pseudo-labels & 0.6859 (+0.0028) & 0.7257 (-0.0122) & 0.8487 (-0.0085) & 0.0922 (-0.0094) & \textbf{0.0585} (-0.0028) \\
ApolloScape & SAM2$\rightarrow$Synthetic & \textbf{0.6940} (+0.0109) & 0.7397 (+0.0017) & 0.8555 (-0.0016) & 0.1046 (+0.0029) & 0.0643 (+0.0030) \\
ApolloScape & Synthetic$\rightarrow$SAM2 & 0.6906 (+0.0075) & 0.7394 (+0.0015) & 0.8536 (-0.0035) & \textbf{0.0916} (-0.0100) & 0.0587 (-0.0026) \\

\midrule
Cityscapes & Target-only baseline & 0.6832 & 0.7377 & 0.8576 & 0.1022 & 0.0616 \\
Cityscapes & source$\rightarrow$target only & 0.5468 (-0.1364) & 0.4839 (-0.2538) & 0.5047 (-0.3529) & 0.0884 (-0.0138) & \textbf{0.0524} (-0.0092) \\
Cityscapes & + Synthetic images & 0.7038 (+0.0206) & \textbf{0.7577} (+0.0200) & \textbf{0.8622} (+0.0046) & 0.0858 (-0.0164) & 0.0552 (-0.0064) \\
Cityscapes & + SAM2 pseudo-labels & 0.6958 (+0.0126) & 0.7371 (-0.0006) & 0.8537 (-0.0040) & 0.0857 (-0.0165) & 0.0545 (-0.0071) \\
Cityscapes & SAM2$\rightarrow$Synthetic & \textbf{0.7064} (+0.0232) & 0.7567 (+0.0190) & 0.8617 (+0.0041) & 0.0915 (-0.0108) & 0.0569 (-0.0047) \\
Cityscapes & Synthetic$\rightarrow$SAM2 & 0.7034 (+0.0202) & 0.7530 (+0.0153) & 0.8588 (+0.0012) & \textbf{0.0808} (-0.0214) & \textbf{0.0524} (-0.0092) \\

\midrule
SANPO & Target-only baseline & 0.6830 & 0.7382 & 0.8572 & 0.1017 & 0.0614 \\
SANPO & source$\rightarrow$target only & 0.5162 (-0.1667) & 0.5716 (-0.1666) & 0.7958 (-0.0614) & 0.2980 (+0.1963) & 0.1599 (+0.0986) \\
SANPO & + Synthetic images & 0.6887 (+0.0058) & \textbf{0.7581} (+0.0199) & \textbf{0.8635} (+0.0063) & 0.0939 (-0.0078) & 0.0580 (-0.0034) \\
SANPO & + SAM2 pseudo-labels & 0.6848 (+0.0019) & 0.7387 (+0.0005) & 0.8572 (+0.0001) & 0.0848 (-0.0169) & 0.0541 (-0.0073) \\
SANPO & SAM2$\rightarrow$Synthetic & \textbf{0.6905} (+0.0075) & 0.7489 (+0.0107) & 0.8613 (+0.0042) & 0.0954 (-0.0063) & 0.0589 (-0.0025) \\
SANPO & Synthetic$\rightarrow$SAM2 & 0.6889 (+0.0059) & 0.7560 (+0.0178) & 0.8614 (+0.0042) & \textbf{0.0830} (-0.0187) & \textbf{0.0527} (-0.0087) \\

\midrule
SideGuide & Target-only baseline & \textbf{0.6829} & 0.7381 & 0.8570 & 0.1018 & 0.0614 \\
SideGuide & source$\rightarrow$target only & 0.4057 (-0.2772) & 0.5312 (-0.2069) & 0.7637 (-0.0934) & 0.2125 (+0.1107) & 0.1199 (+0.0585) \\
SideGuide & + Synthetic images & 0.6805 (-0.0024) & \textbf{0.7543} (+0.0162) & \textbf{0.8592} (+0.0021) & 0.0975 (-0.0043) & 0.0615 (+0.0002) \\
SideGuide & + SAM2 pseudo-labels & 0.6727 (-0.0102) & 0.7259 (-0.0122) & 0.8488 (-0.0082) & 0.0966 (-0.0052) & 0.0605 (-0.0009) \\
SideGuide & SAM2$\rightarrow$Synthetic & 0.6783 (-0.0046) & 0.7356 (-0.0025) & 0.8546 (-0.0025) & 0.1023 (+0.0004) & 0.0634 (+0.0021) \\
SideGuide & Synthetic$\rightarrow$SAM2 & 0.6798 (-0.0031) & 0.7483 (+0.0102) & 0.8562 (-0.0008) & \textbf{0.0902} (-0.0116) & \textbf{0.0582} (-0.0032) \\

\bottomrule
\end{tabular}
}
\end{table*}

\subsection{Offline Accuracy and Safety-Oriented Road/Sidewalk Errors}

Table~\ref{tab:offline_safety_summary} compares the selected offline candidates, representing four model-selection priorities: maximum accuracy, mobile feasibility, balanced accuracy-safety behavior, and architecture-diversity. The accuracy-oriented UPerNet-MobileNetV3 achieved the highest mIoU \((0.715 \pm 0.006)\), but its Road-as-Sidewalk Error Rate remained 0.097. Thus, the best global segmentation score did not correspond to the lowest false-safe road-sidewalk error.

DeepLabV3Plus-MobileNetV3 showed the strongest safety-oriented behavior among the selected candidates. Although its mIoU was lower \((0.690 \pm 0.005)\), it achieved the lowest Road-as-Sidewalk Error Rate \((0.079)\), the highest Road IoU \((0.761)\), and a low Critical False Safe Rate \((0.054)\). The balanced UPerNet-MobileNetV3 candidate retained near-maximum mIoU \((0.714 \pm 0.004)\) while reducing Road-as-Sidewalk Error Rate to 0.084 and matching the lowest Critical False Safe Rate \((0.054)\). SegFormer-MiT-B0 served as the architecture-diversity candidate and had high Sidewalk IoU \((0.864)\), but had the highest Road-as-Sidewalk Error Rate \((0.099)\) and Critical False Safe Rate \((0.063)\).

\begin{figure}[!t]
\centering
\includegraphics[width=0.98\textwidth]{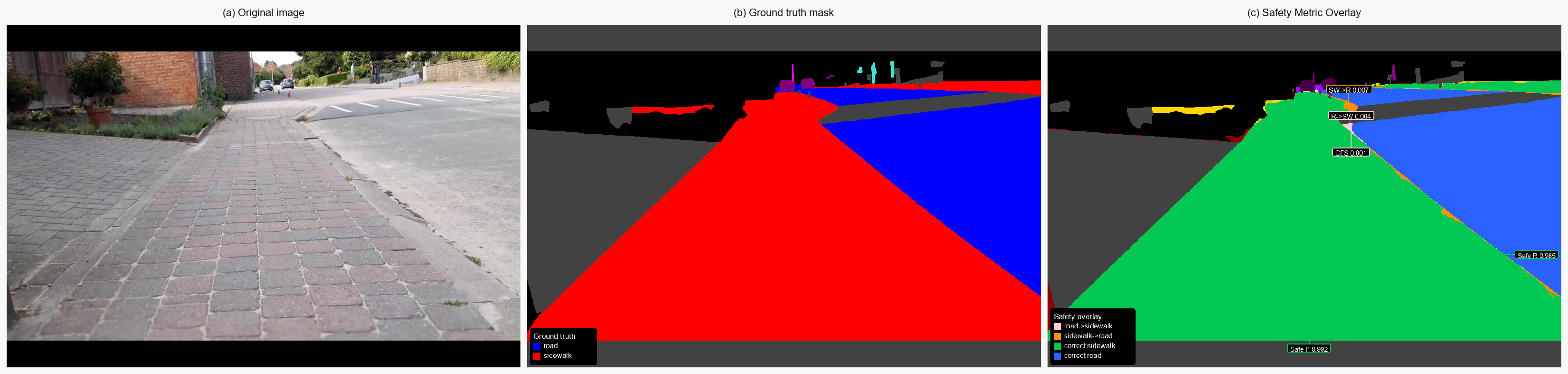}
\vspace{0.5em}

\includegraphics[width=0.98\textwidth]{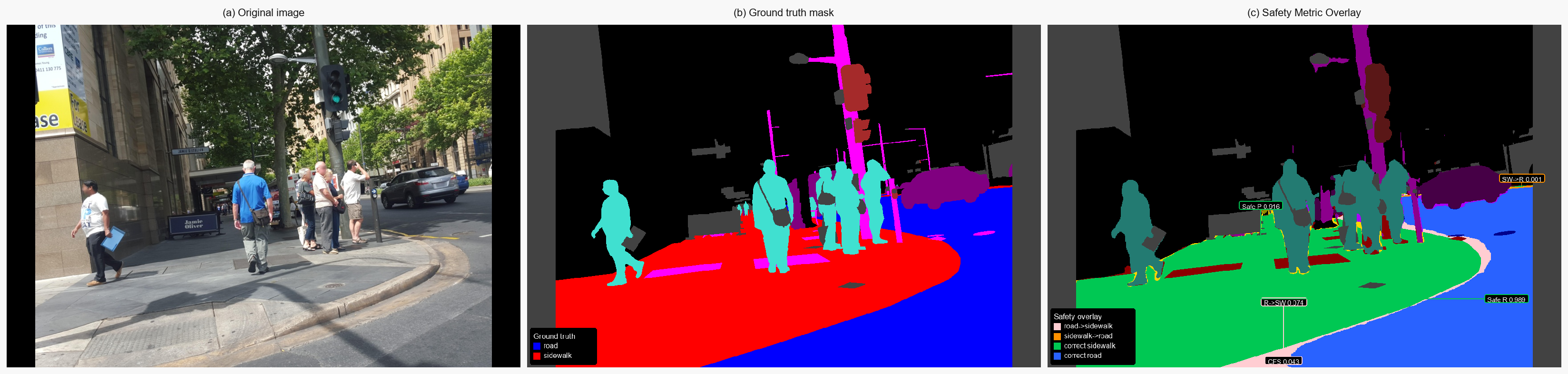}
\vspace{0.5em}

\includegraphics[width=0.98\textwidth]{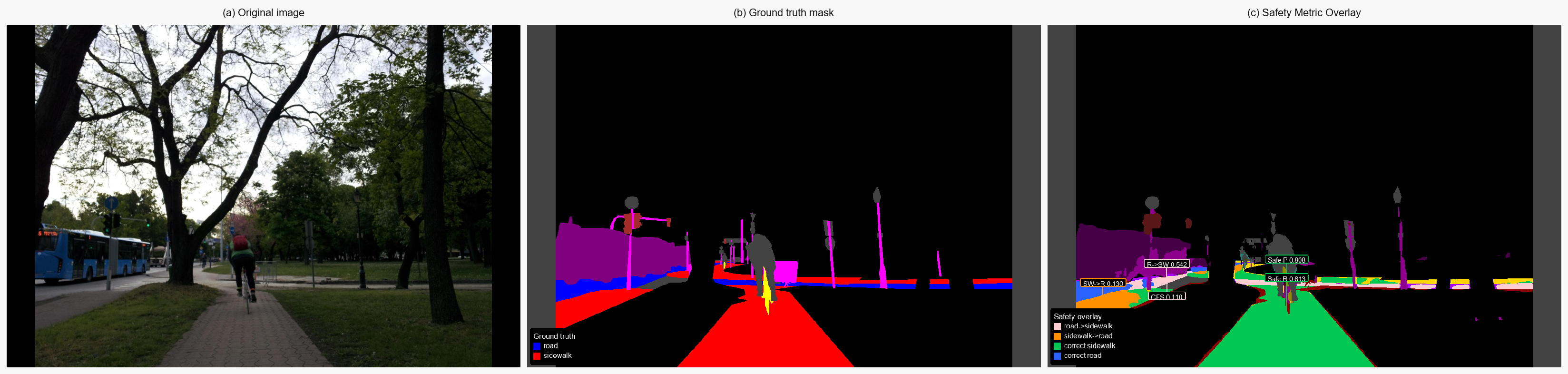}
\caption{\textbf{Qualitative examples of road/sidewalk safety-metric overlays.} Each row shows the original image, ground-truth mask, and safety-metric overlay. The examples illustrate low-error, boundary-error, and severe false-safe cases. Pink indicates road pixels predicted as sidewalk, orange indicates sidewalk pixels predicted as road, green indicates correctly predicted sidewalk, and blue indicates correctly predicted road. The callouts report per-image safety measures, including Road-as-Sidewalk Error Rate, Sidewalk-as-Road Error Rate, Safe-Walkable Precision, Safe-Walkable Recall, and Critical False Safe Rate.}
\altText{Three-row qualitative figure explaining road/sidewalk safety metrics. Each row contains three panels: original camera image, ground-truth semantic mask, and safety-metric overlay. In the top row, the original image is a sidewalk beside a road; the mask shows a large red sidewalk region and a blue road region; the overlay is mostly green for correctly predicted sidewalk and blue for correctly predicted road, with small orange and pink areas near the boundary. This illustrates a low-error case with very limited false-safe Road-as-Sidewalk prediction. In the middle row, the original image shows a sidewalk running beside a road and railing in a built environment; the ground truth separates the road and sidewalk clearly; the overlay contains a visible pink strip along the road-sidewalk boundary, showing road pixels predicted as sidewalk, plus green sidewalk and blue road regions. This illustrates a boundary-error case where false-safe errors occur near the transition between safe and unsafe surfaces. In the bottom row, the original image shows a tree-lined sidewalk scene with a bus and cyclist; the ground-truth mask includes road and sidewalk regions with many narrow objects; the overlay contains extensive pink and orange areas around the road-sidewalk boundary and distant scene structure, indicating substantial confusion. This severe case shows why false-safe Road-as-Sidewalk failure cases need separate evaluation: they can make unsafe road pixels appear walkable. The overlay legend uses pink for road predicted as sidewalk, orange for sidewalk predicted as road, green for correct sidewalk, and blue for correct road.}
\label{fig:safety_metric_overlay}
\end{figure}

Figure~\ref{fig:safety_metric_overlay} illustrates how these metrics correspond to visible road-sidewalk failures. The examples show low-error, boundary-error, and severe false-safe cases, highlighting that Road-as-Sidewalk failure cases can occur locally at road-sidewalk boundaries or more extensively in visually complex scenes.
For assistive sidewalk navigation, candidate selection therefore requires joint interpretation of global segmentation accuracy and road-sidewalk-specific safety metrics.

\begin{table}[!t]
\centering
\caption{\textbf{Offline safety-oriented segmentation summary for selected candidates.} R$\rightarrow$S denotes Road-as-Sidewalk Error Rate, and CFS denotes Critical False Safe Rate. Higher is better for mIoU, Road IoU, and Sidewalk IoU; lower is better for R$\rightarrow$S and CFS. Bold values denote the best-performing result for each metric within this selected-candidate comparison group.}
\altText{Table comparing four selected offline segmentation candidates across mIoU, Road IoU, Sidewalk IoU, Road-as-Sidewalk Error Rate, and Critical False Safe Rate. Higher values are better for mIoU and IoU columns; lower values are better for Road-as-Sidewalk Error Rate and Critical False Safe Rate. Accuracy UPerNet-MobileNetV3 has the highest mIoU, 0.715 plus or minus 0.006, Road IoU 0.760, tied highest Sidewalk IoU 0.864, Road-as-Sidewalk Error Rate 0.097, and Critical False Safe Rate 0.058. DeepLabV3Plus-MobileNetV3 has lower mIoU, 0.690 plus or minus 0.005, but the highest Road IoU 0.761, Sidewalk IoU 0.863, the lowest Road-as-Sidewalk Error Rate 0.079, and tied lowest Critical False Safe Rate 0.054. Balanced UPerNet-MobileNetV3 has mIoU 0.714 plus or minus 0.004, Road IoU 0.760, Sidewalk IoU 0.860, Road-as-Sidewalk Error Rate 0.084, and tied lowest Critical False Safe Rate 0.054. SegFormer-MiT-B0 has mIoU 0.699 plus or minus 0.008, Road IoU 0.752, tied highest Sidewalk IoU 0.864, the highest Road-as-Sidewalk Error Rate 0.099, and the highest Critical False Safe Rate 0.063. The table shows that the highest mIoU candidate is not the safest by false-safe road-sidewalk behavior, while DeepLabV3Plus-MobileNetV3 is strongest on the main safety-error metric.}
\label{tab:offline_safety_summary}
\scriptsize
\setlength{\tabcolsep}{2.5pt}
\begin{tabular}{@{}lccccc@{}}
\toprule
Candidate & mIoU $\uparrow$ & Road IoU $\uparrow$ & Sidewalk IoU $\uparrow$ & R$\rightarrow$S $\downarrow$ & CFS $\downarrow$ \\
\midrule
Accuracy UPerNet-MobileNetV3
& \(\mathbf{0.715 \pm 0.006}\) & 0.760 & \textbf{0.864} & 0.097 & 0.058 \\
DeepLabV3Plus-MobileNetV3
& \(0.690 \pm 0.005\) & \textbf{0.761} & 0.863 & \textbf{0.079} & \textbf{0.054} \\
Balanced UPerNet-MobileNetV3
& \(0.714 \pm 0.004\) & 0.760 & 0.860 & 0.084 & \textbf{0.054} \\
SegFormer-MiT-B0
& \(0.699 \pm 0.008\) & 0.752 & \textbf{0.864} & 0.099 & 0.063 \\
\bottomrule
\end{tabular}
\end{table}

\subsection{Selection of Candidates for Android ONNX Benchmarking}

Android ONNX candidates were selected to preserve the main offline trade-offs rather than only the highest-mIoU model. As summarized in Table~\ref{tab:onnx_candidates}, the selected set includes an accuracy-oriented UPerNet-MobileNetV3 reference, a safety- and deployment-oriented DeepLabV3Plus-MobileNetV3 candidate, a balanced UPerNet-MobileNetV3 order-effect comparison, and a compact transformer-based SegFormer-MiT-B0 candidate. This selection supports smartphone-focused assistive-navigation evaluation by linking offline accuracy, false-safe road-sidewalk behavior, and mobile feasibility.

\begin{table}[!t]
\centering
\caption{\textbf{Selected Android ONNX benchmark candidates.} Candidates were chosen to represent complementary offline model-selection priorities: accuracy, safety-oriented behavior, Stage~3 order comparison, and architecture-diversity.}
\altText{Table listing four Android ONNX benchmark candidates. C1 is UPerNet-MobileNetV3 pretrained on Cityscapes and selected as the accuracy-oriented reference. C2 is DeepLabV3Plus-MobileNetV3 pretrained on SideGuide and selected as the safety- and deployment-oriented candidate because of its low Road-as-Sidewalk Error Rate and mobile-suitable architecture. C3 is UPerNet-MobileNetV3 pretrained on Cityscapes and selected for accuracy-safety trade-off and Stage~3 order comparison. C4 is SegFormer-MiT-B0 pretrained on AllData and selected as a compact transformer-based architecture-diversity candidate.}
\label{tab:onnx_candidates}
\small
\setlength{\tabcolsep}{4pt}
\begin{tabular}{@{}p{0.07\textwidth}p{0.23\textwidth}p{0.15\textwidth}p{0.24\textwidth}p{0.23\textwidth}@{}}
\toprule
ID & Model & Pretraining & Stage & Selection rationale \\
\midrule
C1 & UPerNet-MobileNetV3 & Cityscapes & Stage~3 SAM2$\rightarrow$Synthetic & Accuracy-oriented reference with highest offline mIoU \\
C2 & DeepLabV3Plus-MobileNetV3 & SideGuide & Stage~3 Synthetic$\rightarrow$SAM2 & Safety- and deployment-oriented candidate with low Road-as-Sidewalk Error Rate and mobile-suitable design \\
C3 & UPerNet-MobileNetV3 & Cityscapes & Stage~3 Synthetic$\rightarrow$SAM2 & Accuracy-safety trade-off candidate and UPerNet-MobileNetV3 order-effect comparison \\
C4 & SegFormer-MiT-B0 & AllData & Stage~3 SAM2$\rightarrow$Synthetic & Architecture-diversity candidate using a compact transformer encoder \\
\bottomrule
\end{tabular}
\end{table}

\subsection{Android ONNX Deployment Feasibility}

Table~\ref{tab:android_benchmark_summary} summarizes the Android ONNX benchmark results. Increasing input resolution from 512\(\times\)384 to 768\(\times\)576 generally improved IoU but reduced FPS, confirming the expected accuracy-runtime trade-off for on-device segmentation.

DeepLabV3Plus-MobileNetV3 achieved the highest runtime at both resolutions. At 512\(\times\)384, C2 reached 7.383 FPS with an IoU of 0.723, making it the fastest evaluated operating point. At 768\(\times\)576, its IoU increased to 0.748, while FPS decreased to 3.215. SegFormer-MiT-B0 achieved the highest Android IoU at 768\(\times\)576 \((0.757)\), but with a lower runtime of 1.508 FPS. The UPerNet-MobileNetV3 candidates showed competitive IoU but substantially lower FPS.

When interpreted together with the offline safety metrics, DeepLabV3Plus-MobileNetV3 at 512\(\times\)384 provides the most practical measured operating point for smartphone-based prototype development. These results indicate deployment feasibility under benchmark conditions, but they do not constitute evidence of real-world navigation safety.

\begin{table}[!t]
\centering
\caption{\textbf{Device-balanced Android ONNX benchmark summary.} The table reports frames per second (FPS) and Intersection over Union (IoU). Bold values denote the best result within each input-resolution comparison group.}
\altText{Table summarizing Android ONNX deployment performance for four candidates at two input resolutions. Columns are candidate ID, role, model, resolution, mean FPS, and mean IoU. C1, the UPerNet-MobileNetV3 accuracy reference, reaches 2.792 FPS and the best 512 by 384 IoU of 0.737 at 512 by 384, and 1.245 FPS and 0.744 IoU at 768 by 576. C2, the DeepLabV3Plus-MobileNetV3 deployment and safety-oriented candidate, reaches the best FPS at both resolutions: 7.383 FPS and 0.723 IoU at 512 by 384, and 3.215 FPS and 0.748 IoU at 768 by 576. C3, the UPerNet-MobileNetV3 accuracy-safety trade-off candidate, reaches 2.886 FPS and 0.711 IoU at 512 by 384, and 1.262 FPS and 0.735 IoU at 768 by 576. C4, the SegFormer-MiT-B0 architecture-diversity candidate, reaches 3.842 FPS and 0.733 IoU at 512 by 384, and 1.508 FPS and the best 768 by 576 IoU of 0.757 at 768 by 576. The table shows that higher resolution generally improves IoU but reduces FPS, and that C2 at 512 by 384 is the fastest evaluated operating point.}
\label{tab:android_benchmark_summary}
\small
\setlength{\tabcolsep}{4pt}
\begin{tabular}{@{}p{0.07\textwidth}p{0.20\textwidth}p{0.24\textwidth}p{0.12\textwidth}cc@{}}
\toprule
ID & Role & Model & Resolution & FPS mean $\uparrow$ & IoU mean $\uparrow$ \\
\midrule
C1 & Accuracy reference & UPerNet-MobileNetV3 & 512\(\times\)384 & 2.792 & \textbf{0.737} \\
C1 & Accuracy reference & UPerNet-MobileNetV3 & 768\(\times\)576 & 1.245 & 0.744 \\
C2 & Practical candidate & DeepLabV3Plus-MobileNetV3 & 512\(\times\)384 & \textbf{7.383} & 0.723 \\
C2 & Practical candidate & DeepLabV3Plus-MobileNetV3 & 768\(\times\)576 & \textbf{3.215} & 0.748 \\
C3 & Accuracy-safety trade-off & UPerNet-MobileNetV3 & 512\(\times\)384 & 2.886 & 0.711 \\
C3 & Accuracy-safety trade-off & UPerNet-MobileNetV3 & 768\(\times\)576 & 1.262 & 0.735 \\
C4 & Architecture-diversity & SegFormer-MiT-B0 & 512\(\times\)384 & 3.842 & 0.733 \\
C4 & Architecture-diversity & SegFormer-MiT-B0 & 768\(\times\)576 & 1.508 & \textbf{0.757} \\
\bottomrule
\end{tabular}
\end{table}

\section{Discussion}

This study investigated semantic segmentation as a perception component for smartphone-based assistive navigation for BVIPs. Rather than supporting autonomous navigation or route planning, the intended role is to identify walkable sidewalk regions, estimate road-boundary proximity, and detect obstacles or uncertain road-sidewalk boundaries that may later be communicated through user-facing feedback.

A central finding is that model selection for this application cannot rely on mIoU alone. Although mIoU measures overall segmentation performance, it does not indicate whether unsafe road regions are predicted as walkable sidewalk. We therefore evaluated Road-as-Sidewalk Error Rate and Critical False Safe Rate as proxy safety measures to complement conventional segmentation metrics. The offline comparison showed that the highest-mIoU model was not the safest, demonstrating that segmentation accuracy and safety-oriented behavior should be considered jointly.

The staged target-domain adaptation results further showed that direct source-to-target transfer was insufficient for pedestrian-view sidewalk segmentation, highlighting the need for target-domain adaptation. Synthetic images primarily improved segmentation accuracy and road/sidewalk IoU, whereas SAM2 pseudo-labels more consistently reduced Road-as-Sidewalk Error Rate. Their complementary behavior suggests that both approaches are valuable as controlled training-stage extensions rather than replacements for target-domain data.

The Android ONNX benchmark linked these offline findings to practical deployment. DeepLabV3Plus-MobileNetV3 achieved the best measured on-device runtime while maintaining favorable safety-oriented behavior, demonstrating that lightweight semantic segmentation models can support smartphone-based assistive perception.

The reported safety measures remain proxy metrics derived from offline segmentation masks rather than validated indicators of real-world navigation safety. Consequently, future user studies are required to determine how segmentation outputs should be translated into effective guidance and whether they improve mobility, user confidence, and trust. The involvement of a blind first author helped ensure that dataset design, class selection, and interpretation of segmentation errors reflected mobility-relevant priorities.

Overall, the findings support evaluating assistive sidewalk segmentation using three complementary perspectives: segmentation accuracy, safety-oriented road-sidewalk behavior, and practical smartphone deployment.

\FloatBarrier

\section{Conclusions}

This study presented a safety-oriented semantic segmentation framework for smartphone-based assistive navigation for BVIPs. Using the proposed SENSATION-DS dataset and a unified pedestrian-view taxonomy, we evaluated transfer learning, staged target-domain adaptation, controlled synthetic augmentation, SAM2-generated pseudo-labels, and smartphone deployment under a common safety-oriented evaluation framework.

The results demonstrate that overall segmentation accuracy alone is insufficient for selecting models intended for assistive navigation. Synthetic augmentation primarily improved segmentation accuracy, whereas SAM2 pseudo-labels more consistently reduced Road-as-Sidewalk false-safe errors. Among the evaluated candidates, UPerNet-MobileNetV3 achieved the highest offline mIoU, while DeepLabV3Plus-MobileNetV3 provided the strongest combination of safety-oriented behavior and measured smartphone performance.

Future work will expand SENSATION-DS with more diverse pedestrian environments, improve quality control of synthetic and pseudo-labeled supervision, and integrate the selected segmentation model into a complete smartphone or wearable navigation prototype for evaluation with BVIP users. Ultimately, real-world user studies are required to validate whether improved segmentation and reduced false-safe errors translate into safer and more effective assistive navigation.

\section*{Acknowledgements}

The authors gratefully acknowledge the scientific support and HPC resources provided by the Erlangen National High Performance Computing Center (NHR@FAU) of the Friedrich-Alexander-Universit\"at Erlangen-N\"urnberg (FAU). The hardware is partially funded by the German Research Foundation (DFG).

\section*{Funding}

No funding was received for this work.

\section*{Conflict of Interest / Disclosure Statement}

The authors declare no conflicts of interest.

\section*{Data \& Code Availability Statement}

Due to privacy constraints, raw images containing public-space scenes are available only upon reasonable request and subject to anonymization and institutional approval. Aggregated benchmark tables, trained ONNX model metadata, and evaluation scripts can be made available in a public repository.

\section*{Author Contribution Statement}

Hakan Calim and Anamaria Dumitrescu contributed to the conceptualization, methodology, software development, experimentation, analysis, and drafting of the manuscript. Adarsh Bhandary Panambur coordinated the work, contributed to the methodology and validation, and managed and revised the manuscript. Huzaifa Asif conducted the Android deployment and model testing. Andreas Maier provided senior supervision, contributed to the conceptualization, methodology, and analysis, provided resources, and critically revised the manuscript. All authors analyzed the results, contributed expertise through extensive discussions, and reviewed and approved the final manuscript.

\section*{Generative AI Use Statement}

OpenAI ChatGPT (GPT-5.5) was used for language editing and improving the clarity of selected manuscript passages. All AI-assisted text was reviewed and approved by the authors, who take full responsibility for the manuscript.

\bibliographystyle{unsrtnat}
\bibliography{references}
\end{document}